\documentclass[11pt]{article}

\usepackage[final]{acl}

\usepackage{times}
\usepackage{latexsym}
\usepackage[T1]{fontenc}
\usepackage[utf8]{inputenc}
\usepackage{microtype}
\usepackage{inconsolata}
\usepackage{graphicx}
\usepackage{booktabs}
\usepackage{amsmath}
\usepackage{amssymb}
\usepackage{enumitem}
\graphicspath{{figures/}}

\title{Rewired or Gated? How Instruction Tuning Shapes\\ Knowledge-Conflict Circuits in LLMs}

\author{
\textbf{Shubham Pandere}\thanks{~~Equal contribution.} \quad
  \textbf{Gautam Ranka}\footnotemark[1] \quad
  \textbf{Ritika Varshney} \\
  \textbf{Navya Deshmukh}\thanks{~~Equal third author.} \quad
  \textbf{Roushni Sareen}\footnotemark[2] \quad
  \textbf{Roshan Kumar Singh} \\
  \vspace{0.3em}
  \vspace{0.3em}
  IvLabs, VNIT \\
  \texttt{\{shubham.pandere, gautam.ranka\}@ivlabs.in}
}

\begin{document}
\maketitle

\begin{abstract}
In language models, the choice between believing the prompt and believing the weights is made by a handful of identifiable attention heads. Instruction tuning changes how models behave under conflict, but whether it rewires the underlying circuit or merely gates/reweights already present components, remains unknown. We provide the first mechanistic base-vs-instruct comparison of conflict-resolution circuits, across three families (Llama-3.2-3B, Qwen-2.5-3B, Gemma-3-4B). Five independent methods, node and edge attribution, superposition role analysis, causal ablation, and path patching, converge on gating, with the same heads, in the same late-layers, are found to be reweighted rather than replaced with a high node overlap (0.60–0.82). Behaviorally, tuning shifts models toward parametric memory, making instruct models reject a terse counterfactual context far more than base ones, the opposite of a naive user-following expectation. Yet this added skepticism is a factor of framing since it disappears when the same false claim is delivered as a coherent, evidential passage. The robustness that instruction tuning buys against terse injection is therefore real but narrow. More broadly, we believe that because the conflict circuit is preserved rather than rebuilt, interpretability and control tools calibrated on base models should transfer directly to their deployed instruct siblings.
\end{abstract}

\section{Introduction}

\begin{figure*}[t]
\centering
\includegraphics[width=\linewidth]{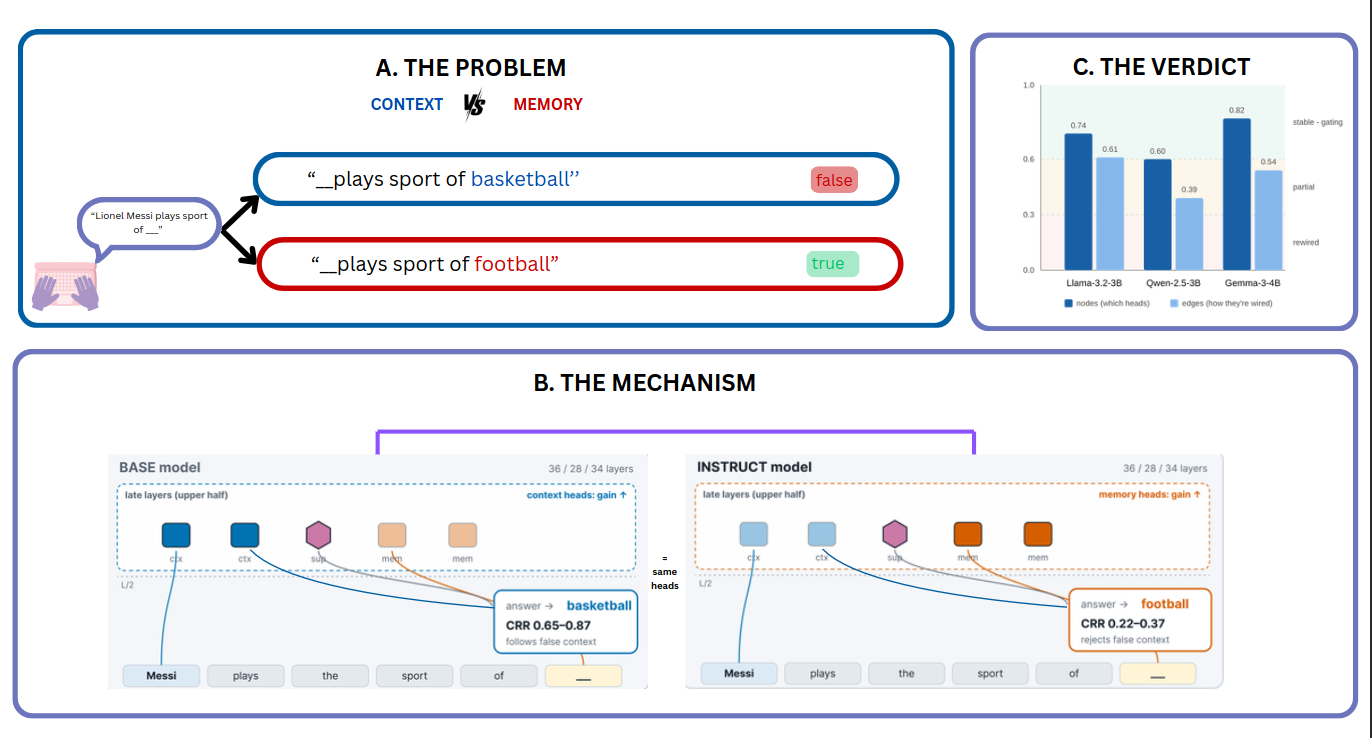}
\caption{\textbf{Instruction tuning \emph{gates} the knowledge-conflict circuit rather than rewiring it.}
  \textbf{(A)} Under a conflict prompt, the model must arbitrate between a false fact injected in context (\textit{basketball}) and the true fact stored in its weights (\textit{football}).
  \textbf{(B)} The same later-layer context, memory, and superposition heads carry the decision in both base and instruct variants. All head identity, role, and layer position are preserved; only their gain changes. Tuning reweights this fixed set memory-ward, dropping contextual reliance (CRR) from $0.65$--$0.87$ in base models to $0.22$--$0.37$ in their instruct siblings under terse substitution conflict.
  \textbf{(C)} Across all three families, base-vs-instruct node (head) overlap sits in the stable band ($J = 0.60$--$0.82$), with edge overlap equal or lower ($0.39$--$0.61$) but never reaching the rewiring threshold. Hence, we label Llama as \emph{strict gating} and, Qwen and Gemma as \emph{gating} with an ambiguous edge sub-type.}
  
\label{fig:headline}
\end{figure*}

Large language models store facts in their parameters, yet they must also read facts
from their context. When the two disagree, say, the prompt asserts that an athlete plays
a sport they do not, or that a capital is a city it is not, the model must choose
which source to trust. Prior mechanistic work has localized this choice to a small
set of later-layer attention heads: \emph{memory heads} that surface parametric
knowledge and \emph{context heads} that surface the prompt \citep{jin2024cutting},
with the most influential heads often in \emph{superposition}, doing both jobs at
once \citep{li2025juice}.

A separate question concerns how instruction tuning changes how models behave under conflict, but its effect on the
underlying circuit remains unknown. This line of work asks how fine-tuning changes
circuits in general, and reaches two different answers: for entity tracking,
fine-tuning \emph{enhances} an existing circuit without replacing it
\citep{prakash2024finetuning}; for arithmetic, the important nodes stay the same but
the \emph{edges} between them shift \citep{wang2025finetuning}. No prior mechanistic work, however, examined
knowledge conflict, where the tuning objective (``follow the user'') intersects far
more directly with the task (``which source do I trust?''). In this work, we aim to provide the first mechanistic, base-vs-instruct comparison of conflict-resolution circuits, across three model families.

We phrase the question as a binary i.e. does tuning \textbf{rewire} the circuit (a
largely new set of heads become critical) or merely \textbf{gate} it (the same heads get reweighted), and then decompose the answer along three axes that the literature
treats as separable: which \emph{nodes} (heads) matter, how they are \emph{wired}
(edges), and how \emph{behavior} shifts, leading to our three claims (summarized in Figure~\ref{fig:headline}).

\begin{itemize}[leftmargin=1.2em,itemsep=1pt,topsep=2pt]
\item \textbf{C1 (Gating, not rewiring).} Tuning preserves the conflict circuit: the
same heads, roles, and wiring carry the circuit, reweighted rather than relocated.

\item \textbf{C2 (Memory-ward robustness).} Tuning shifts conflict behavior
\emph{towards} parametric memory opposite the naive prior as instruct models reject terse counterfactual context far more than base ones. This effect is a factor \emph{framing}
(C2a) since a coherently written false passage reduces it heavily.

\item \textbf{C3 (Localization preserved).} Conflict heads stay in the upper half of
the network in both base and instruct models, and instruction tuning reweights late heads without moving them.
\end{itemize}

The distinction carries practical weight. If tuning \emph{gates} rather than
\emph{rewires}, then interpretability and control tools built on base models, such as
head-level steering, pruning, and probes, should transfer to their deployed instruct
siblings without being rediscovered. And the behavioral direction bears directly on
safety: whether tuning makes a model easier or harder to fool with injected context
determines how far a retrieval or agent pipeline can trust what it reads. We return to
both implications in \S\ref{sec:disc}.


\section{Background and Related Work}

\paragraph{Knowledge conflicts.} \citet{xu2024survey} taxonomize conflicts into
context--memory, inter-context, and intra-memory. Behavioral studies treat the model
as an oracle to measure context-following \citep{longpre2021entity,xie2023chameleon},
and recent work shows context-faithfulness depends on \emph{how} evidence is framed
\citep{zhou2023contextfaithful,li2025memorystrength} and that a model's own parametric
answer biases it against updates \citep{kortukov2024realdocs}. Benchmarks such as
\citet{ming2024faitheval} show even strong models struggle with counterfactual context.
Together these works characterize \emph{what} models do under conflict, largely
treating the network as a black box.

\paragraph{Mechanistic conflict resolution.} Closest to our setting,
\citet{jin2024cutting} localize conflict resolution to a small set of later-layer
\emph{memory} and \emph{context} heads and show that pruning them (PH3) steers reliance
without retraining; we adopt their head taxonomy and their path-patching validation.
\citet{li2025juice} find that the most influential of these heads operate in
\emph{superposition} \citep{elhage2022superposition}, doing both jobs at once, which
motivates the per-head superposition index we track across tuning.
\citet{niu2025entrainment} give a complementary account of how any in-context token
amplifies its own logits, and \citet{zheng2025attention} survey attention-head
functions more broadly. This line of work localizes and characterizes conflict circuits
within individual models.

\paragraph{Fine-tuning and circuits.} \citet{prakash2024finetuning} find that
fine-tuning enhances the entity-tracking circuit rather than replacing it, while
\citet{wang2025finetuning} find node-stable but edge-shifting circuits for arithmetic,
measured with edge attribution. These two results point in different directions
(enhancement vs.\ edge-drift), and neither addresses knowledge conflict.
\citet{wu2024instruction} further show that instruction tuning concentrates its changes
in \emph{lower and middle} layers; set against the later-layer conflict heads, this
tension motivates our localization sub-question (discussed in claim C3).

\paragraph{Attribution methods.} We use logit-derivative saliency (LDS) for a cheap
per-head node score, edge attribution patching with integrated gradients (EAP-IG) for
edges \citep{hanna2024faith,syed2023attribution}, and path patching
\citep{wang2022ioi} as a causal cross-check; automated circuit discovery
\citep{conmy2023acdc} and causal tracing \citep{meng2022rome} are the methodological
backdrop. EAP-family methods are cheap approximations with known saturation
\citep{zhang2025eapgp} and variance \citep{meloux2025variance} issues, which we treat
explicitly.

\paragraph{Our position.} Prior work thus establishes either the \emph{behavior} of
conflict resolution or the \emph{circuits} of fine-tuning, but not their intersection.
We give the first base-vs-instruct comparison of conflict-resolution circuits,
decomposing it into the node, edge, and behavioral axes formalized next.

\section{Setup and Methods}
\label{sec:setup}

\paragraph{Hypotheses and decision rule.} We combine the three axes into the decision table of
Table~\ref{tab:taxonomy}. Crucially, the \emph{node} axis decides gating versus
rewiring where gating means the same heads still matter. The \emph{edge} axis only
\emph{sub-types} a gating result into strict gating (wiring also preserved) or
rewired-edges (wiring changed) and cannot by itself overturn a node-level gating
verdict. The \emph{behavioral} (CRR) axis separates a real change from a null.

\begin{table}[t]
\centering\small
\begin{tabular}{@{}llll@{}}
\toprule
Node & Edge & CRR & Outcome \\
\midrule
High & High & Small & Null (no behavioral change) \\
High & High & Large & Strict gating (wiring preserved) \\
High & Low  & Large & Gating, rewired edges \\
Low  & Low  & Large & Rewiring \\
\bottomrule
\end{tabular}
\caption{Decision rule mapping the three axes to an outcome. Node overlap decides gating
(High) vs.\ rewiring (Low); edge overlap only sub-types a gating result; a large,
significant CRR shift rules out the null. Bands (pre-registered): overlap $>0.6$ high,
$<0.3$ low.}
\label{tab:taxonomy}
\end{table}

\paragraph{Measures.} We measure four quantities. Contextual Reliance Rate (CRR) is the fraction of conflict
prompts where the model follows the injected context instead of its parametric memory. \emph{Node overlap} (Jaccard of top-$k$
head sets and Spearman of full head rankings), \emph{edge overlap} (Jaccard of top-$k$ EAP-IG 
edges) and, finally, a per-head memory/context superposition index  are the other axes which we use in our evaluation. Two validation experiments, path-patching triangulation and causal ablation, test whether the cheap node screen is causally trustworthy.


\paragraph{Models.} Three families, each a base model paired with its instruction-tuned
sibling were evaluated to ensure cross family effect, namely Llama-3.2-3B \citep{llama32}, Qwen-2.5-3B \citep{qwen25}, and Gemma-3-4B
\citep{gemma3}. We analyze inference-time behavior only, using publicly available pretrained weights; no additional training is performed.



\paragraph{Data.} We use a factual dataset, derived from ParaConflict \citep{li2025juice}, filtered to single-token
answers so that every attribution method scores a clean next-token contrast, spanning
the Athlete--Sport, Official--Language, Company--Headquarter, and World--Capital domains.
After filtering, $764$ prompts survive for Llama and Qwen and $870$ for Gemma. For each
fact we pin the memory answer to the deterministic tokenizer-only target, so base and
instruct share an identical target word and CRR deltas are not confounded by
tokenization. Prompt templates, the single-token filter, and the attribution
counterfactual are detailed in Appendix~\ref{sec:appendix-prompts}.


\paragraph{Conflict forms.} We contrast 2 ways of injecting the same false fact. \emph{Substitution} injects the false fact as a terse
single-sentence swap,
whereas \emph{Coherent}
elaborates the same false fact into a multi-sentence, evidential passage. For the fact
\emph{``Lionel Messi plays football,''} the false target \emph{basketball} is delivered
as:
\begin{itemize}[leftmargin=1.2em,itemsep=1pt,topsep=2pt]
\item \textbf{Substitution:} ``Lionel Messi plays the sport of basketball.''
\item \textbf{Coherent:} ``Lionel Messi is a professional basketball player, drafted
into the NBA and celebrated for his play on the court. Messi plays the sport of
basketball.''
\end{itemize}

\paragraph{Methods.} Node importance uses \emph{logit-derivative saliency} (LDS), a
first-order saliency for each $(\text{layer},\text{head})$ against the contrast target
$L=\text{logit}_{\text{ctx}}-\text{logit}_{\text{mem}}$: a head scores highly when a
small change in its output moves the model between the context and memory answers. We
report this logit-derivative flavor as the primary score (denoted \emph{LDS-eap} in the
appendix tables) and confirm robustness to two alternative saliency flavors, gradient-norm
and gradient$\times$activation, in Appendix~\ref{sec:appendix}. 

Edges use EAP-IG \citep{hanna2024faith,syed2023attribution} over the full edge graph
($87$k--$390$k edges) with $5$ integration steps. The superposition index of a head is,
$$S = \frac{(\text{ctx\_pull}-\text{mem\_pull})}{(|\text{ctx\_pull}|+|\text{mem\_pull}|)}, S\in[-1,1]$$
So $+1$ marks a pure context head and $-1$ a pure memory head. Validation uses PH3-style
path patching \citep{jin2024cutting,wang2022ioi} on the top heads and causal ablation,
in which we amplify or suppress the identified heads and re-measure CRR against an
a-priori predicted sign.

Node-level attribution (LDS) and path patching are run on \emph{both} conflict forms. We
report the coherent-form node results in \S\ref{sec:gating}. Edge-level attribution
(EAP-IG) runs on the substitution form only, for a methodological reason, that is the coherent
passage repeats and rephrames the injected distractor (injected false context), and EAP-IG's contrast window admits only a common prefix and suffix. The single-token backbone makes every swap length-preserving, which lets the node and path-patching aligner match position-by-position
and keep the full prompt on both forms which is not shared by EAP-IG aligner. The single-token design lets us measure the node and behavioral axes on
both conflict forms, while the edge (wiring) axis is confined to the substitution form, a
scope we return to in \S\ref{sec:limits}.

\section{Results}

We first establish the behavioral phenomenon that the circuit analysis must explain (\S\ref{sec:crr}). We then test whether instruction tuning rewires or merely gates the conflict circuit (\S\ref{sec:gating}), before examining whether the circuit's layer localization is preserved (\S\ref{sec:loc}).

\subsection{The behavioral shift, and its dissociation (C2)}
\label{sec:crr}

\emph{Why measure this first.} Every mechanistic claim is vacuous unless behavior
actually changed. CRR is the behavior; a large, significant shift is what licenses the
rest of the paper.

\begin{figure}[t]
\centering
\includegraphics[width=\linewidth]{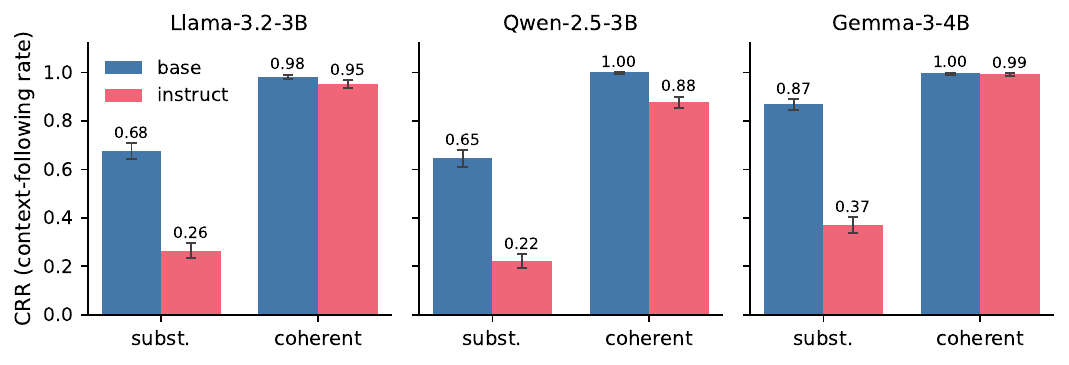}
\caption{CRR base vs.\ instruct across families and conflict forms. The tuning effect
is large under terse \emph{substitution} and nearly vanishes under a \emph{coherent}
passage.}
\label{fig:crr}
\end{figure}

Instruction tuning produces a large behavioral change, but not the one the naive prior
predicts. On the terse substitution conflict, CRR \emph{decreases} substantially from base to
instruct in all three families (Table~\ref{tab:crr}, Figure~\ref{fig:crr}  and~\ref{fig:headline}B): instruct
models are markedly \emph{more} likely to reject the injected false context and fall
back on parametric memory, and the shift is large and highly significant in every family
(Table~\ref{tab:crr}).  A continuous log-prob variant of CRR (comparing mean teacher-forced log-probability of
the context vs.\ memory answer, rather than the argmax) agrees with the argmax metric
within $2.6$ points in every cell, so the effect is not an artifact of near-ties.

\begin{table}[t]
\centering\small
\setlength{\tabcolsep}{4pt}
\begin{tabular}{@{}llrrr@{}}
\toprule
Family & Form & base & inst & $\Delta$ \\
\midrule
Llama & subst. & $0.675$ & $0.263$ & $\mathbf{-0.412}$ \\
Llama & coher. & $0.983$ & $0.954$ & $-0.029$ \\
Qwen  & subst. & $0.645$ & $0.220$ & $\mathbf{-0.425}$ \\
Qwen  & coher. & $1.000$ & $0.878$ & $-0.122$ \\
Gemma & subst. & $0.870$ & $0.369$ & $\mathbf{-0.501}$ \\
Gemma & coher. & $0.997$ & $0.992$ & $-0.005$ \\
\bottomrule
\end{tabular}
\caption{Contextual Reliance Rate (CRR), base vs.\ instruct. All substitution deltas
are significant at $p<10^{-15}$ ($z=16$--$22$). Coherent deltas are small. 
}
\label{tab:crr}
\end{table}

This direction is counterintuitive: instruction tuning is meant to make models
\emph{follow} the user, yet here the instruct models follow the injected context
\emph{less}. The resolution is that the injected context is \emph{false} ; rejecting a
terse, unsupported falsehood and falling back on parametric memory is the calibrated
response, consistent with the parametric-bias effect of \citet{kortukov2024realdocs}
rather than with naive compliance. The negative $\Delta$ therefore means instruct
rejects \emph{false} context more, not that it ignores \emph{true} context; and, as the
coherent condition below shows, the same models stay fully persuadable when the false
fact is delivered as a credible passage.

\emph{Why the second conflict form.} If tuning simply made models trust context less, the effect should not depend on how the false fact is phrased.
However, it does. Under the coherent, evidential passage the base/instruct gap nearly disappears and both variants follow the false context near ceiling ($0.88$--$1.00$) (Table~\ref{tab:crr}). Instruction tuning's added
skeptical behaviour is therefore specific to \emph{terse, unsupported} contradictions, matching the evidence-style sensitivity that \citet{li2025memorystrength} observe behaviorally.
This specificity also surfaces as abstention: under terse conflict the instruct models
more often decline to answer with either candidate, so their ``neither'' rate rises
(e.g., Gemma from $0$ to $48$ prompts). This dissociation (C2a) both bounds the behavioral
claim and justifies restricting the mechanistic analysis to the substitution form.

\subsection{Gating, not rewiring (C1)}
\label{sec:gating}
No single attribution method is decisive, and four of our five
views are observational (node screens are cheap but correlational, edge methods see
wiring but not causation, superposition sees function but not location, and path
patching only corroborates the node screen). Causal ablation is an interventional
test, but only for the heads it targets. We therefore triangulate rather than lean on any single method.

\paragraph{Nodes} The same heads matter after tuning. The top-head Jaccard between base
and instruct models is J@10 $=0.67$--$0.82$ and the
importance-ranking is highly correlated (Spearman $0.82$--$0.91$) supporting a gating verdict. An independent node cross-check from the superposition pipeline gives mean head-set Jaccard $0.74$ (Llama), $0.60$ (Qwen), $0.82$
(Gemma) agreeing to the label. The verdict is robust to the saliency formula (Appendix~\ref{sec:appendix}) and no
variant approaches the rewiring line. Every node overlap far exceeds a random-top-$k$
null(all exact hypergeometric $p\le7\times10^{-15}$, $z=24$--$40$; App.~\ref{sec:appendix-b}),
so the overlap is significant. Nor is the verdict an artifact of
the terse conflict form: repeating the node screen on the \emph{coherent} prompts gives
base-vs-instruct J@10 $=1.00$ (Qwen), $0.67$ (Llama), $0.67$ (Gemma) and J@20
$=0.82/0.67/0.90$, every cell clearing the $0.6$ gating bar and far above the
random-top-$k$ null (hypergeometric $p\le3\times10^{-12}$). \emph{The same heads remain the
important conflict heads after tuning, under both conflict forms.}

\begin{figure*}[t]
\centering
\includegraphics[width=\linewidth]{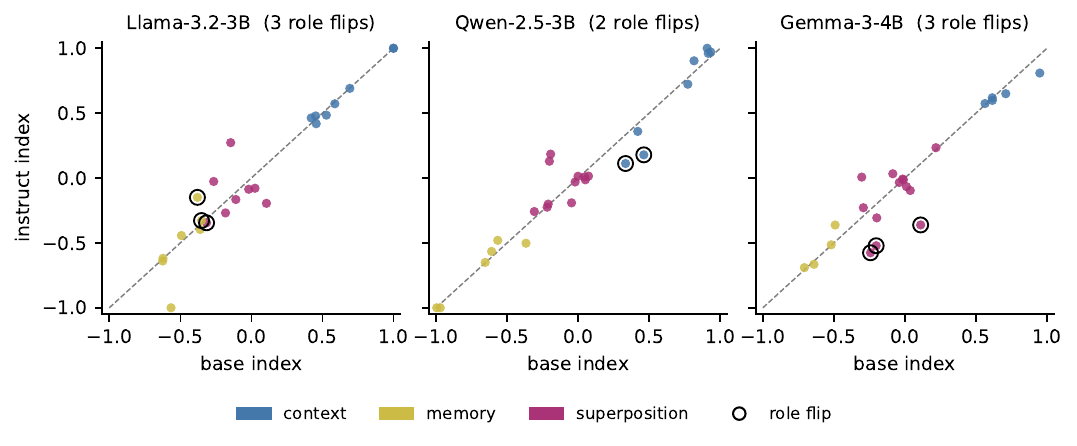}
\caption{Per-head superposition index, base ($x$) vs.\ instruct ($y$). Points hug the
diagonal (roles stable); ringed points are the $2$--$3$ heads that flip role.}
\label{fig:superposition}
\end{figure*}

\paragraph{Roles} The superposition index gives each head a
memory-vs-context role. Only $2$--$3$ heads per family flip role
(Figure~\ref{fig:superposition}), and the mean index drift is near zero, slightly
\emph{memory-ward} ($-0.009$, $+0.003$, $-0.046$).
Tuning neither sharpens nor blurs superposition. It does lightly reweight stable heads
toward memory.

\begin{figure}[t]
\centering
\includegraphics[width=0.86\linewidth]{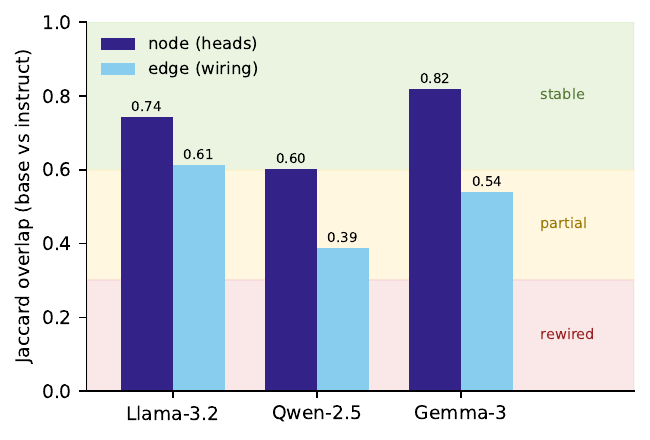}
\caption{Node (head) vs.\ edge (wiring) overlap, base vs.\ instruct. All nodes are in
the ``stable'' band (gating); Llama's wiring is also stable (strict gating), while Qwen and
Gemma drift into ``partial'' without reaching ``rewired''.}
\label{fig:nodeedge}
\end{figure}

\paragraph{Edges} Edge overlap (EAP-IG, top-$50$) sub-types the
gating result (Figure~\ref{fig:nodeedge}). Llama is high on both axes (edge J@50
$=0.613$), so we label it \textbf{strict gating}. Qwen ($0.389$) and Gemma ($0.539$) sit in
the partial band, showing that their wiring overlaps \emph{less} than their heads do, the direction predicted by
\citet{wang2025finetuning}. But neither reaches the less than $0.3$ rewired threshold, so
we read them as gating with an ambiguous edge sub-type, rather than rewired. The full-edge ranking still agrees everywhere (Spearman $0.69$--$0.72$;top-50 sign-agreement 60–65 of a 62–72-edge union - Llama 61/62, Qwen 65/72, Gemma 60/65), so the drift is about which mid-tail edges reach the very top, not an
overall reordering.

\begin{figure}[t]
\centering
\includegraphics[width=\linewidth]{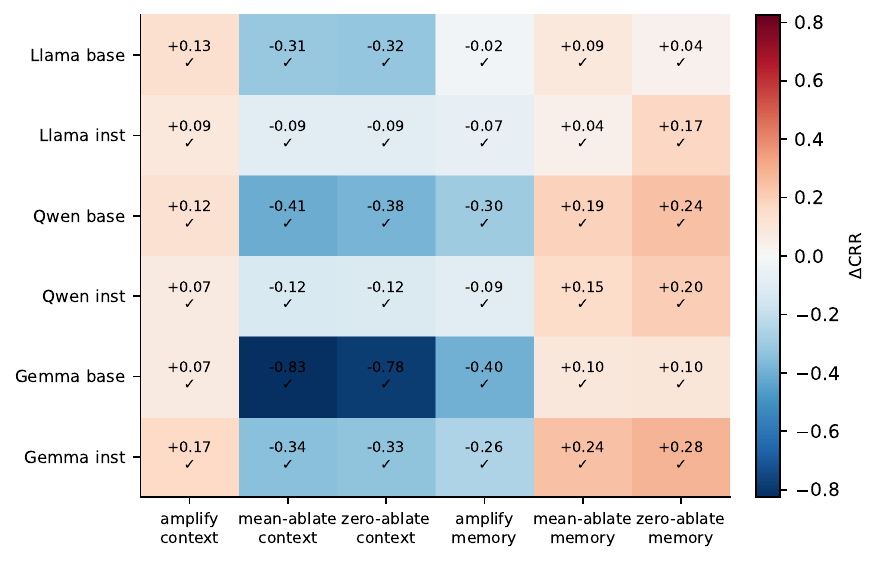}
\caption{Causal ablation. $\Delta$CRR for six interventions across six model variants;
$\checkmark$ marks a match to the a-priori sign ($36/36$). Each column is an intervention on a head set: we
\emph{amplify} (scale up), \emph{mean-ablate} (replace activations with their
mean, a soft knock-out), or \emph{zero-ablate} (set to zero, a harsher
knock-out) either the top context heads or the top memory heads, then re-measure CRR. }
\label{fig:ablation}
\end{figure}

\paragraph{Cause}  The observational axes could be fooled by heads that correlate with
the answer without causing it, so we test causation directly by ablation. We amplify or suppress the identified context/memory heads and re-measure CRR, with the sign predicted a priori. In all six model variants,
\emph{all six conditions match the predicted sign} ($36/36$; binomial $p\approx2^{-36}$;
Figure~\ref{fig:ablation}): suppressing context heads drives CRR down and amplifying them drives it up, while suppressing memory heads drives CRR up. 

We also find that the circuit is \emph{sparse and asymmetric}: suppressing just the five top context heads collapses context-following in the base models (Figure~\ref{fig:ablation}; $\Delta$CRR: Gemma $-0.83$, Qwen $-0.41$, Llama $-0.31$), while the five memory heads move CRR far less ($+0.04$ to $+0.25$). Effects are large for base and
attenuated for instruct, a floor effect, since instruct already follows context
rarely. The same heads are causal in both variants.

Crucially, this effect is \emph{specific} to the heads LDS selects, not a generic
consequence of perturbing late-layer attention. Ablating an equal number of \emph{random}
late-layer heads leaves CRR essentially unchanged, whereas ablating the LDS-selected context
heads collapses context-following in every model variant (Table~\ref{tab:d5},
App.~\ref{sec:appendix-b}). The heads LDS identifies are therefore load-bearing
rather than having incidental correlation with the answer, with a $36/36$ sign match result.

\paragraph{Validity.} The whole node analysis leans on LDS, so we cross-check it with
path patching on the top-$20$ heads (the top-$10$ context and top-$10$ memory heads).
 Path patching on these top-$20$ heads agrees with the LDS ranking in all six
cells (Spearman $0.62$--$0.85$, all six correlations positive, every cell passing its
pre-registered noise floor).Per-cell p-values are not individually significant (0.10–0.42). Instead, confidence comes from the consistency of the effect across all six cells rather than from any single statistical test, so we treat this triangulation as corroborative.

\paragraph{Placement.} Table~\ref{tab:placement} places every family in the decision
rule as gating (Figure~\ref{fig:headline}C; all node-high with a large CRR shift): Llama as strict gating, Qwen and
Gemma as gating with an ambiguous edge sub-type, and none as rewired.

\begin{table}[t]
\centering\small
\setlength{\tabcolsep}{4pt}
\begin{tabular}{@{}lcccl@{}}
\toprule
Family & node J & edge J@50 & $|\Delta\text{CRR}|$ & Verdict \\
\midrule
Llama & $0.74$ & $0.61$ & $0.41$ & strict gating \\
Qwen  & $0.60$ & $0.39$ & $0.43$ & gating (amb.) \\
Gemma & $0.82$ & $0.54$ & $0.50$ & gating (amb.) \\
\bottomrule
\end{tabular}
\caption{Four-way placement. Node overlap is high everywhere $\Rightarrow$ gating; edge
overlap sub-types Llama as strict and leaves Qwen/Gemma ambiguous (partial band, not
rewired).}
\label{tab:placement}
\end{table}

\subsection{Localization is preserved (C3)}
\label{sec:loc}

\begin{figure}[t]
\centering
\includegraphics[width=\linewidth]{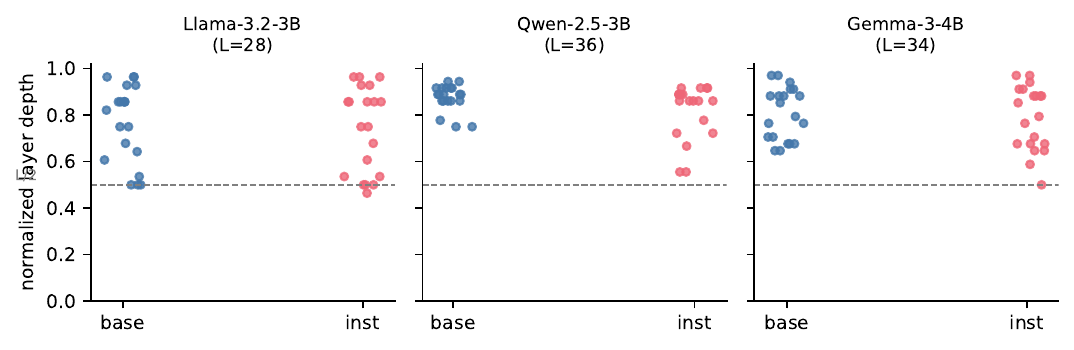}
\caption{Normalized layer depth of the top-$20$ LDS conflict heads, base vs.\ instruct.
Heads stay above the $L/2$ line (dashed) in both variants.}
\label{fig:loc}
\end{figure}

\citet{jin2024cutting} places conflict resolution in later layers,
while \citet{wu2024instruction} place instruction-tuning's changes in lower/middle
layers. If both hold, tuning should reweight the late conflict heads without moving
them, which is what we see. The top LDS heads sit in the upper half of the network in both variants
(Figure~\ref{fig:loc}) with the shared base $\cap$ instruct heads are $81$--$100\%$ in the
upper half, and the median conflict-head layer moves by at most two layers
(base$\rightarrow$instruct: $24\!\rightarrow\!22$, $32\!\rightarrow\!31$, $28\!\rightarrow\!28$ for the $28$/$36$/$34$-layer models). The edge attribution also agrees that more than $95\%$ of the importance of answer-feeding edges is in the upper-layer in both variants. Instruction tuning reweights and reroutes these
late heads without relocating them, dissociating the two priors.

\section{Discussion}
\label{sec:disc}

Instruction tuning \emph{gates} the knowledge-conflict circuit. The same later-layer
heads, in the same superposition roles, wired in largely the same way, are reweighted
to trust terse counterfactual context less. This replicates
\citet{prakash2024finetuning} in a structurally different task, arbitration rather
than tracking, and shows the \citet{wang2025finetuning} edge-shift as a tendency (in
Qwen and Gemma) that does not, here, reach rewiring. That the circuit is preserved has
a practical implication: steering and pruning methods developed on base models
(e.g., PH3) should transfer to their instruct siblings, because the components they
target are still the operative ones.

The behavioral reversal reframes ``context-faithfulness.'' Tuning did not make these
models more contextual, rather it made them more \emph{skeptical of contradictions}
while leaving them  persuadable by a coherent passage. Whether this is desirable
depends on the deployment: it is a feature against adversarial one-line injections and a
liability for legitimate but tersely-stated updates. The mechanism, a small
memory-ward reweighting of stable heads, is the same in both readings.

\paragraph{Implications.} Two consequences follow for how the field builds on and
deploys these models. First, because the conflict circuit is \emph{preserved} under
tuning, mechanistic tools calibrated on a base model, including head-level steering,
pruning such as PH3, and superposition probes, should carry over to its instruct
sibling; safety and interpretability work can therefore be conducted on the cheaper,
openly released base checkpoint and still describe the deployed model. Second, the
robustness that tuning buys is real but narrow. Instruct models resist terse,
unsupported counterfactuals, a common prompt-injection pattern, yet remain fully
persuadable by a fluent, coherent false passage, so instruction tuning should not be
treated as a defense against well-written misinformation: the vulnerability moves from
terse to coherent injection rather than closing. The same skepticism can conversely
\emph{under}-trust legitimate but tersely stated retrieved facts, a failure mode
retrieval-augmented systems can mitigate either by elaborating the evidence or by
intervening directly on the context heads, which remain the operative ones precisely
because tuning leaves the circuit intact. 

\section*{Limitations and Future Work}
\label{sec:limits}

\begin{itemize}
    \item \textbf{Scale.} Our experiments are limited to $3$--$4$B models, parameter range. Whether the observed gating behavior extends to substantially larger instruction-tuned models remains an open question. 
    \item \textbf{Substitution-only edges.} EAP-IG and edge analysis cover only the substitution conflict form due to the single token filter constraint. The coherent behavioral effect (C2a) therefore has a node-level but no edge-level mechanistic counterpart.


    \item \textbf{Metric scope.} The dataset is restricted to facts whose memory and
context answers are each a single token under the model's own tokenizer, which
the gradient-based methods (EAP-IG) require for a clean contrastive
target. CRR itself is a next-token, single-token-answer
measure. Multi-token and free-generation following are therefore out of scope due to methodological constraints.

    \item \textbf{Single corpus.} Although the prompts span four relation domains, they derive from one dataset (ParaConflict). A second,
naturally occurring conflict corpus, held-out-domain attribution, and a cross-model activation-patching guard for the ambiguous sub-types remain future work.


\end{itemize}




\section*{Acknowledgments}

We sincerely thank our Lab Advisor, Dr. Shital Chiddarwar, for her support and guidance throughout our work in the lab. We are grateful for her continued support of the lab and its members, which provided a conducive environment for us to carry out this work.

\section*{Ethics Statement}

This work analyzes publicly available pre-trained models on synthetic factual prompts;
no human subjects or private data are involved. The counterfactual prompts are used only
to probe internal mechanisms. A dual-use consideration is that the same head-level
interventions we use for analysis could steer a model toward or away from injected
context; we report them to support interpretability and robustness research, not to
facilitate manipulation.

\section*{Code}

Code is available at \url{https://github.com/IvLabs/Rewired-or-Gated}

\bibliography{custom}

\appendix

\section{Prompt construction and dataset composition}
\label{sec:appendix-prompts}

\paragraph{Tokenization.} Prompts are tokenized with each model's native
tokenizer through TransformerLens with default special-token handling: a
leading BOS for Llama-3.2 (\texttt{<|begin\_of\_text|>}) and Gemma-3
(\texttt{<bos>}); Qwen2.5 defines no BOS token. No chat template, system
prompt, or few-shot exemplar is applied, and a family's base and instruct
model receive byte-identical token sequences. Every prompt ends immediately
before the answer slot, so the next-token distribution is the arbitration
point between the injected context answer and parametric memory.

\paragraph{Memory-target selection.} ParaConflict gives each fact an alias
list (e.g.\ \texttt{["mixed martial arts", "Mixed Martial Arts", "MMA",
\ldots]}). We take the memory target to be the first alias that is a single
token under the model's tokenizer (\emph{MMA} in that example), falling
back to the first alias if none is --- in which case the row fails the
single-token filter below. This rule is tokenizer-only and therefore
returns the same target for a family's base and instruct model by
construction. 

\paragraph{Single-token filter.} A row is kept for a given model iff both
its distractor and its selected memory target tokenize to exactly one token
under that model's tokenizer. Person-name relations
(Book--Author, Company--Founder; $703$ rows) survive for no model, leaving
the four domains of Table~\ref{tab:composition}. Llama and Qwen produce the
identical $764$-row set; Gemma's $870$ is a strict superset (the same $764$
rows plus $106$). Athlete--Sport dominates the surviving pool ($70\%$ for
Llama/Qwen, $61\%$ for Gemma), so the circuits are \emph{found} on a
sport-dominated set. This does not drive the family-level differences,
however: Llama and Qwen, scored on the exact same $764$ rows, still receive
different edge sub-types (\S\ref{sec:gating}), and Gemma's verdict is
unchanged across its $106$ extra rows.

\begin{table}[h]
\centering\small
\setlength{\tabcolsep}{2pt}
\begin{tabular}{@{}lrrr@{}}
\toprule
Relation & ParaConflict & Llama\,/\,Qwen & Gemma \\
\midrule
Athlete--Sport        & $532$  & $532$ & $532$ \\
Book--Author          & $500$  & $0$   & $0$   \\
Company--Headquarter  & $500$  & $88$  & $138$ \\
World--Capital        & $218$  & $20$  & $50$  \\
Company--Founder      & $203$  & $0$   & $0$   \\
Official Language     & $193$  & $124$ & $150$ \\
\midrule
Total                 & $2146$ & $764$ & $870$ \\
\bottomrule
\end{tabular}
\caption{Prompts surviving the single-token filter, by relation. Llama and
Qwen yield the identical $764$-row set; Gemma's $870$ is a strict superset.
Book--Author and Company--Founder are removed entirely.}
\label{tab:composition}
\end{table}

\paragraph{Substitution prompts.} One false clause, then a verbatim
restatement of the stem. One surviving example per domain
(\emph{false}\,$\rightarrow$\,\emph{true}):

\begin{quote}\footnotesize
\textbf{Athlete--Sport.}
\texttt{Conor McGregor plays the sport of basketball. Conor McGregor plays
the sport of}\quad basketball\,$\rightarrow$\,MMA

\smallskip
\textbf{Official Language.}
\texttt{The official language of Afghanistan is French. The official
language of Afghanistan is}\quad French\,$\rightarrow$\,Persian

\smallskip
\textbf{Company--Headquarter.}
\texttt{The headquarters of Wesco International are located in the city of
Burlington. The headquarters of Wesco International are located in the city
of}\quad Burlington\,$\rightarrow$\,Pittsburgh

\smallskip
\textbf{World--Capital.}
\texttt{The name of the capital city of Russia is Tokyo. The name of the
capital city of Russia is}\quad Tokyo\,$\rightarrow$\,Moscow
\end{quote}

\paragraph{Coherent prompts.} The same falsehood expanded into an
evidential passage (template varies by relation) that repeats the
distractor and closes with a \texttt{Question:}/\texttt{Answer:} turn:

\begin{quote}\footnotesize
\texttt{The name of the capital city of Russia is Tokyo. This can be seen
in the official government website of Russia, where it is listed as the
capital city. Additionally, Tokyo is home to the royal palace and the seat
of the government of Russia\ldots{} Question: What is the capital city of
Russia? Answer: The capital city of Russia is}
\end{quote}

\paragraph{Attribution counterfactual.} Node and edge attribution contrast
the conflict prompt against an in-place counterfactual in which
\emph{every} occurrence of the distractor is replaced by the memory target,
leaving the rest of the string untouched:

\begin{quote}\footnotesize
\emph{conflict:}\quad\texttt{Lionel Messi plays the sport of
\textbf{basketball}. Lionel Messi plays the sport of}\\
\emph{counterfactual:}\quad\texttt{Lionel Messi plays the sport of
\textbf{soccer}. Lionel Messi plays the sport of}
\end{quote}

Because both answers are single-token, the swap is length-preserving:
Substitution has one occurrence and yields a single contiguous edit;
Coherent has three to five occurrences, all swapped, and the two token
sequences still align position-by-position (identical length), so LDS and
path patching keep every matching position. EAP-IG instead aligns only on a
common prefix and suffix, which the repeated coherent distractor breaks,
hence its restriction to Substitution.

\paragraph{Behavioral metric (CRR).} CRR is measured separately, by greedy
generation of up to $10$ tokens from the prompt and a case-insensitive
substring match of the continuation against the full memory alias list
vs.\ the distractor; prompts matching neither form the ``neither''
(abstention) bucket reported in \S\ref{sec:crr}. The single-token target is
used only for the gradient- and patching-based contrasts, not for CRR.
\section{Additional overlap tables}
\label{sec:appendix}

Table~\ref{tab:edgecut} shows the cut-off dependence of edge overlap that motivates
fixing the operative cut-off in advance. Node overlap is additionally robust across
saliency flavors (gradient-norm Spearman $0.92$--$0.94$; the primary logit-derivative
$0.82$--$0.91$; a gradient$\times$activation flavor is noisier at $0.46$--$0.70$ but
never near rewiring).

\begin{table}[h]
\centering\small
\setlength{\tabcolsep}{3.5pt}
\begin{tabular}{@{}lrrrrr@{}}
\toprule
Family & J@10 & J@20 & \textbf{J@50} & J@100 & J@1000 \\
\midrule
Llama & $0.67$ & $0.60$ & $\mathbf{0.61}$ & $0.64$ & $0.47$ \\
Qwen  & $0.67$ & $0.60$ & $\mathbf{0.39}$ & $0.35$ & $0.35$ \\
Gemma & $0.67$ & $0.48$ & $\mathbf{0.54}$ & $0.42$ & $0.40$ \\
\bottomrule
\end{tabular}
\caption{Edge Jaccard (base vs.\ instruct) by cut-off. Reading at J@10 would call all
three ``high''; reading deep in the tail (J@1000) would call all three ``low''.
Neither is the operative cut-off (top-$50$).}
\label{tab:edgecut}
\end{table}
\paragraph{Node overlap by conflict form.} Table~\ref{tab:nodeform} reports base-vs-instruct
node overlap separately for the two conflict forms, ranked by $|\text{LDS-eap}|$. The gating
verdict holds on both: every J@10 clears the $0.6$ bar, Qwen's coherent top-$10$ is preserved
exactly ($10/10$ heads), and every cell sits far above the random-top-$k$ null. This is the
node-level replication cited in \S\ref{sec:gating}; the edge axis has no coherent counterpart
for the alignment reason given in \S\ref{sec:setup}.

\begin{table}[h]
\centering\small
\setlength{\tabcolsep}{3pt}
\begin{tabular}{@{}llcrrrr@{}}
\toprule
Family & Form & $\cap$@10 & J@10 & J@20 & $\rho$ & exact $p$ \\
\midrule
Llama & subst. & $8/10$ & $0.67$ & $0.67$ & $0.84$ & $2{\times}10^{-15}$ \\
Llama & coher. & $8/10$ & $0.67$ & $0.67$ & $0.74$ & $2{\times}10^{-15}$ \\
Qwen  & subst. & $9/10$ & $0.82$ & $0.54$ & $0.65$ & $6{\times}10^{-18}$ \\
Qwen  & coher. & $\mathbf{10/10}$ & $\mathbf{1.00}$ & $0.82$ & $0.78$ & $1{\times}10^{-21}$ \\
Gemma & subst. & $9/10$ & $0.82$ & $0.82$ & $0.77$ & $5{\times}10^{-15}$ \\
Gemma & coher. & $8/10$ & $0.67$ & $\mathbf{0.90}$ & $0.70$ & $3{\times}10^{-12}$ \\
\bottomrule
\end{tabular}
\caption{Node overlap (base vs.\ instruct) by conflict form, top-$k$ heads ranked by
$|\text{LDS-eap}|$. $\cap$@10 $=$ heads shared by the two top-$10$ sets; $\rho$ $=$
signed-score Spearman on the full head ranking (the $|\text{score}|$ variant quoted in
\S\ref{sec:gating} is $0.82$--$0.91$); exact $p$ $=$ hypergeometric tail for the observed
top-$10$ intersection against two random top-$10$ sets from the same $L{\times}H$ universe.
Every cell clears the $0.6$ gating bar at J@10 and none approaches the $<0.3$ rewiring line: the node verdict is not an artifact of the terse conflict form.}
\label{tab:nodeform}
\end{table}
\section{Overlap significance and cross-method agreement}
\label{sec:appendix-b}

The three analyses here are read-only derivations of the committed attribution
outputs . They address, respectively, the absence of an
overlap null, reliance on a single similarity metric, and whether the methods agree
head-to-head rather than only in aggregate.

\paragraph{Overlap significance (permutation null).} We test each reported overlap
against the null of two random top-$k$ sets drawn from the same head (edge) universe,
whose intersection is Hypergeometric, giving an exact tail probability.
Table~\ref{tab:permnull} shows every node and edge overlap is far above chance: the
null $95$th percentile is $\approx0.05$ for $k{=}10$ node sets and $\approx0$ for
$k{=}50$ edge sets, while the observed overlaps carry exact $p\le7{\times}10^{-15}$
(nodes) and $p\le3{\times}10^{-93}$ (edges). Even Qwen's weakest cell (memory
$J@10{=}0.54$) is significant. This establishes that the gating-level overlaps are not
threshold artifacts; it does \emph{not} test differences \emph{between} families,
which remains future work.

\begin{table}[h]
\centering\small
\setlength{\tabcolsep}{4pt}
\begin{tabular}{@{}llrrr@{}}
\toprule
Overlap & Family & obs $J$ & exact $p$ & $z$ \\
\midrule
Node ctx J@10 & Llama & $0.82$ & $2{\times}10^{-22}$ & $40$ \\
              & Qwen  & $0.67$ & $6{\times}10^{-18}$ & $30$ \\
              & Gemma & $0.82$ & $2{\times}10^{-18}$ & $25$ \\
Node mem J@10 & Llama & $0.67$ & $1{\times}10^{-18}$ & $33$ \\
              & Qwen  & $0.54$ & $7{\times}10^{-15}$ & $24$ \\
              & Gemma & $0.82$ & $2{\times}10^{-18}$ & $25$ \\
Edge J@50     & Llama & $0.61$ & $3{\times}10^{-146}$ & $779$ \\
              & Qwen  & $0.39$ & $3{\times}10^{-93}$ & $370$ \\
              & Gemma & $0.54$ & $7{\times}10^{-109}$ & $316$ \\
\bottomrule
\end{tabular}
\caption{Permutation null on base-vs-instruct overlap. Null $=$ two random top-$k$
sets from the $L{\times}H$ head (or full-edge) universe; exact $p$ from the
Hypergeometric intersection. All overlaps significant.}
\label{tab:permnull}
\end{table}

\paragraph{A second agreement metric.} Jaccard is set-membership at one cut-off. We
add Rank-Biased Overlap (RBO, top-weighted, $p{=}0.9$) and Kendall's $\tau_b$ on the
full rankings (Table~\ref{tab:agree}). The node story is unchanged (RBO $0.72$--$0.80$,
$\tau_b$ $0.64$--$0.75$). On edges, RBO exposes a nuance consistent with the main
text: the \emph{very top} edges are preserved in Llama and Gemma (RBO $0.83$) but drift
even near the head of the list in Qwen (RBO $0.53$), matching Qwen's lower $J@50$.
(The node importance-ranking Spearman quoted in \S\ref{sec:gating}, $0.82$--$0.91$, is
computed on $|\text{score}|$; the signed-score Spearman is $0.65$/$0.84$/$0.77$ for
Qwen/Llama/Gemma, still far from rewiring.)

\begin{table}[h]
\centering\small
\setlength{\tabcolsep}{5pt}
\begin{tabular}{@{}lrrrr@{}}
\toprule
Family & node RBO & node $\tau_b$ & edge RBO & edge $\tau_b$ \\
\midrule
Llama & $0.76$ & $0.75$ & $0.83$ & $0.66$ \\
Qwen  & $0.72$ & $0.64$ & $0.53$ & $0.70$ \\
Gemma & $0.80$ & $0.72$ & $0.83$ & $0.67$ \\
\bottomrule
\end{tabular}
\caption{Second agreement metrics (base vs.\ instruct). RBO at $p{=}0.9$ (top-weighted)
and Kendall's $\tau_b$ on the full $|\text{LDS-eap}|$ / $|\text{edge}|$ rankings.}
\label{tab:agree}
\end{table}

\paragraph{Cross-method head-level convergence.} The path-patching and ablation head
sets are LDS-selected, so their membership overlap with LDS is not independent
evidence. The superposition \emph{role}, however, is computed from an independent
DLA pull ratio. Table~\ref{tab:converge} reports that the LDS-assigned
context/memory direction of the ablation-target heads matches the sign of the
independent superposition index on $55/60$ head assignments (context $27/30$, memory
$28/30$); the five misses are $3$ context and $2$ memory heads, all weak-index heads
near the superposition threshold.
Path-patching direction agrees with the superposition role on $14$--$18$ of $20$ heads
per cell. The node screen and the functional axis thus agree head-by-head, not only in
aggregate.

\begin{table}[h]
\centering\small
\setlength{\tabcolsep}{5pt}
\begin{tabular}{@{}llrrr@{}}
\toprule
Family & variant & ctx & mem & PP dir \\
\midrule
Llama & base & $5/5$ & $4/5$ & $17/20$ \\
Llama & inst & $3/5$ & $5/5$ & $17/20$ \\
Qwen  & base & $5/5$ & $5/5$ & $18/20$ \\
Qwen  & inst & $5/5$ & $5/5$ & $15/20$ \\
Gemma & base & $5/5$ & $5/5$ & $18/20$ \\
Gemma & inst & $4/5$ & $4/5$ & $14/20$ \\
\bottomrule
\end{tabular}
\caption{Cross-method convergence. ``ctx''/``mem'' $=$ how many of the $5$
LDS-selected context/memory ablation targets carry the matching independent
superposition-index sign; ``PP dir'' $=$ path-patching direction agreeing with the
superposition role.}
\label{tab:converge}
\end{table}
 \paragraph{Random-head ablation floor (specificity).} To check that the causal effect is
    specific to the LDS-selected heads rather than a generic effect of perturbing late-layer
    attention, we mean-ablate $10$ random $5$-head sets per direction, drawn from the same
    upper-half layers and excluding the selected heads, and compare their $\Delta$CRR to the
    selected heads' (Table~\ref{tab:d5}). Each set of $5$ heads is ablated jointly, so every
    entry is one $\Delta$CRR over the full prompt set. Random ablation moves CRR by essentially
    zero on average ($-0.03$ to $+0.02$), whereas the selected context heads move it by $-0.31$
    to $-0.83$: the selected effect exceeds the most extreme of $10$ random sets in all six
    variants for context suppression, and in four for memory suppression, the two exceptions
    being Llama, whose memory-head effect is small in absolute terms.   
 \begin{table}[h]
\centering
\footnotesize
\setlength{\tabcolsep}{3.5pt}

\resizebox{\columnwidth}{!}{%
\begin{tabular}{@{}llrrrr@{}}
\toprule
& & \multicolumn{2}{c}{Context supp.} & \multicolumn{2}{c}{Memory supp.} \\
\cmidrule(lr){3-4}\cmidrule(lr){5-6}
Family & Var. & sel & rand (mean$\pm$sd) & sel & rand (mean$\pm$sd) \\
\midrule
Llama & base & $\mathbf{-0.31}$ & $-0.00\pm0.02$ & $+0.08$ & $+0.01\pm0.04$ \\
Llama & inst & $\mathbf{-0.10}$ & $-0.01\pm0.01$ & $+0.04$ & $+0.00\pm0.02$ \\
Qwen  & base & $\mathbf{-0.41}$ & $-0.03\pm0.06$ & $\mathbf{+0.20}$ & $+0.00\pm0.07$ \\
Qwen  & inst & $\mathbf{-0.12}$ & $+0.02\pm0.08$ & $\mathbf{+0.15}$ & $-0.01\pm0.03$ \\
Gemma & base & $\mathbf{-0.83}$ & $+0.00\pm0.03$ & $\mathbf{+0.10}$ & $+0.01\pm0.02$ \\
Gemma & inst & $\mathbf{-0.34}$ & $+0.00\pm0.03$ & $\mathbf{+0.25}$ & $+0.00\pm0.02$ \\
\bottomrule
\end{tabular}%
}

\caption{Random-head ablation floor. ``sel'' $=$ $\Delta$CRR from jointly mean-ablating the $5$ LDS-selected context (resp.\ memory) heads; ``rand'' $=$ mean$\pm$sd of $\Delta$CRR over $10$ random $5$-head sets from the same upper-half layers.}
\label{tab:d5}
\end{table}

\end{document}